\documentclass[10pt,twocolumn,letterpaper]{article}

\usepackage{3dv}
\usepackage{times}
\usepackage{epsfig}
\usepackage{graphicx}
\usepackage{amsmath}
\usepackage{amssymb}
\usepackage{comment}
\usepackage{amsmath,amssymb} %
\usepackage{color}
\usepackage[bottom]{footmisc}
\usepackage{booktabs} 
\usepackage{caption} 
\usepackage{mathtools}
\usepackage[pagebackref=true,breaklinks=true,letterpaper=true,colorlinks,bookmarks=false]{hyperref}

\threedvfinalcopy %

\def\threedvPaperID{395} %

\ifthreedvfinal\fi
\begin{document}

\title{ \vspace{-1.1cm} Intrinsic Autoencoders for Joint Deferred Neural Rendering  \\
and Intrinsic Image Decomposition \vspace{-0.5cm}}

\author{
Hassan Abu Alhaija$^{*,}$\textsuperscript{1}\
\hspace{2mm}
Siva Karthik Mustikovela$^{*,}$\textsuperscript{1}\
\hspace{2mm}
Justus Thies\textsuperscript{2} \\
Varun Jampani\textsuperscript{3} \hspace{2mm}  \
Matthias Nie{\ss}ner\textsuperscript{2} \hspace{2mm} \
Andreas Geiger\textsuperscript{4,5} \hspace{2mm} \
Carsten Rother\textsuperscript{1} \\ 
\\
\textsuperscript{1}Heidelberg University \qquad \
\textsuperscript{2}Technical University Munich \qquad \
\textsuperscript{3}Google Research \qquad \ 
\\
\textsuperscript{4}Max Planck Institute for Intelligent Systems, T\"ubingen \qquad \
\textsuperscript{5}University of T\"ubingen \qquad \
\\ 
}

\maketitle

\begin{abstract}
\vspace{-0.3cm}
Neural rendering techniques promise efficient photo-realistic image synthesis while providing rich control over scene parameters by learning the physical image formation process.
While several supervised methods have been proposed for this task, acquiring a dataset of images with accurately aligned 3D models is very difficult.
The main contribution of this work is to lift this restriction by training a neural rendering algorithm from unpaired data.
We propose an autoencoder for joint generation of realistic images from synthetic 3D models while simultaneously decomposing real images into their intrinsic shape and appearance properties. 
In contrast to a traditional graphics pipeline, our approach does not require to specify all scene properties, such as material parameters and lighting by hand.
Instead, we learn photo-realistic deferred rendering from a small set of 3D models and a larger set of unaligned real images, both of which are easy to acquire in practice.
Simultaneously, we obtain accurate intrinsic decompositions of real images while not requiring paired ground truth.
Our experiments confirm that a joint treatment of rendering and decomposition is indeed beneficial and that our approach outperforms state-of-the-art image-to-image translation baselines both qualitatively and quantitatively.
\end{abstract}

\section{Introduction}

\begin{figure}[ht!]
\centering
\includegraphics[width=\linewidth]{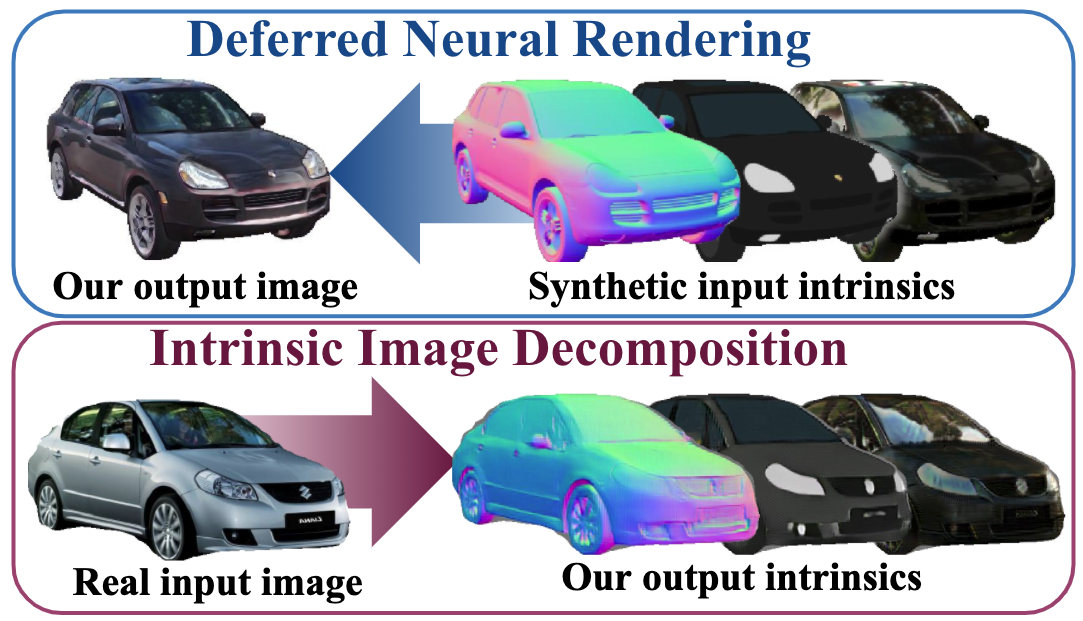}
\caption{\textbf{Deferred Neural Rendering and Intrinsic Image Decomposition.}
At training time, our model exploits normals, albedo and reflections from a small set of 3D models as well as a large set of {\it unpaired} RGB images of the same object category. Our model solves two tasks simultaneously: (i) generating photo-realistic images given the input geometry and basic intrinsic properties, and (ii) decomposing real images back into their intrinsic components.}
\label{fig:intro:teaser}
\end{figure}

State-of-the-art sampling-based rendering engines (e.g., Mitsuba\cite{Mitsuba}) are able to generate photo-realistic images of virtual objects which are nearly indistinguishable from real-world photographs. 
However, this is not an easy task to accomplish since all intrinsic physical aspects of the virtual object must be accurately modeled, such as accurate 3D geometry, detailed textures and physically-based materials. 
While some of these intrinsics are abundant on the internet, such as the geometry of 3D objects (e.g. Turbosquid and 3D Warehouse), others are hard to obtain, such as high-quality materials -- ideally in the form of a highly-accurate spatially-varying BRDF. 
In addition, sophisticated and slow rendering algorithms with many tunable parameters (lighting, environment map, camera model, post-processing) are required for turning 3D content into photo-realistic images.
These parameters are often tuned individually with each rendered image, making it hard to create a large and diverse set of rendered images.
On the other hand, obtaining a large number of real images which capture the complex interaction of light with scene geometry and surface properties is easy.
This makes the idea of learning neural image synthesis from real images very attractive.

Several works on conditional image generation \cite{spade2019,pix2pixhd2018wang,pix2pix2017isola,crn2017chen} have exploited paired datasets of real images with semantic information, including semantic segmentation \cite{spade2019,crn2017chen} and body part labels \cite{Lassner:GeneratingPeople:2017} for training realistic image synthesis models.   
However, such sparse inputs only allow limited control over the generated image. 
This limits the applicability of these methods, e.g., in virtual reality or video game simulations where precise control over the output is essential.
Training a conditional image generation model from richer control inputs would require a large dataset of paired real images with pixel aligned intrinsic properties such as 3D structure, textures, materials and reflections.
Obtaining such a dataset is hard in practice.

Our goal is to take a step towards learning a highly controllable realistic image synthesis model without requiring real world images with aligned 3D models. 
Our key insight is that learning the inverse task of intrinsic decomposition is helpful for learning image synthesis from real images and vice-versa.
We therefore train both, the forward rendering process and the reverse intrinsic decomposition process, jointly using a single objective as illustrated in Fig.~\ref{fig:intro:teaser}.
Inspired by recent results in unpaired image-to-image translation \cite{munit2018huang,unit2017liu,cyclegan2017zhu}, we train our model using an small set of synthetic 3D models of an object category as well as a large unpaired dataset of real images of the same category.   

Towards this goal, we exploit a technique from real-time rendering called \textit{Deferred Rendering} which splits the rendering process into two stages and thus improves efficiency.
In the first stage, the geometry of the scene along with its textures and material properties are projected onto a 2D pixel grid, resulting in a set of 2D intrinsic images which capture the geometry and appearance of the object.
This step is efficient since it does not require physically accurate path tracing but relies on simple rendering operations.
In the second ``deferred'' stage, lighting, shading and textural details are added to form the final rendered image.
Our goal is to replace this second deferred stage of the rendering process with a neural network which we call \textit{Deferred Neural Rendering} (DNR) network.  
To ensure that the input information is represented in the output image, we decompose it back into its intrinsics using a second Intrisic Image Decomposition (IID) network. 
However, we found that using this cycle alone leads to overfitting, especially in the IID network. 
To improve the IID network, we introduce a second \textit{Decomposition cycle}
in which we train the IID network to decompose real images.

Overall, our model follows a similar dual cycle training setup as proposed in \cite{cyclegan2017zhu} and \cite{dualgan}. 
However, an important conceptual difference to these works is that our task is not a one-to-one but a one-to-many mapping.
Different realistic images can be generated from the same set of intrinsic maps as the intrinsics do not uniquely define the image.
Likewise, a single image can be explained using different intrinsic decompositions due to projection from the higher dimensional intrinsics into the RGB image space. 

We therefore introduce a shared adversarial discriminator between the input and the reconstruction at the end of each cycle.
Our model enables both highly photo-realistic image synthesis and accurate intrinsic image decomposition.
We summarize our main contributions as follows:
\begin{itemize}
    \renewcommand{\labelitemi}{\textbullet}
    \item We propose the Intrinsic Autoencoder, a method to jointly train photo-realistic image synthesis and intrinsic image decomposition using cycle consistency losses without using any paired data.
    \item We propose a shared discriminator network that enables better generalization and proves key for learning both tasks without paired training data.
    \item We analyze the importance of various model components using quantitative metrics and human experiments. We also show that our method recovers accurate intrinsic maps from challenging real images.
\end{itemize}

\vspace{-0.2cm}
\section{Related Work}
\noindent \textbf{Differentiable Rendering.}
A standard way of synthesizing images from a given geometry and material is to use rendering engines. 
Several works try to implement the rendering process in a differentiable manner, amenable to neural networks. 
The work of~\cite{blanz1999morphable} used differentiable rendering with
deformable face models for face reconstruction. The works of~\cite{loper2014opendr} and ~\cite{kato2018neural} proposed rasterization-based
differentiable renderers but only support local illumination. In order to support more realistic image formation, some other works~\cite{che2018inverse,gkioulekas2016evaluation,gkioulekas2013inverse,li2018differentiable} propose to back-propagate though path tracing. Differentiable rasterizers are relatively fast, but at the same time highly restrictive as they do not support complex global illumination. While differentiable path tracers produce more realistic images, they are usually quite slow, thus restricting their usage to specific applications. Another drawback of differentiable renderers is that they require a detailed representation of the rendering input in terms of geometry, illumination, materials and viewpoint. In this work, we bypass the specification of complex image formation by training a CNN to directly generate realistic images from given geometry and material inputs. 

\vspace{-0.2mm}
\noindent \textbf{Neural Image Synthesis.}
Generative models such as Generative Adversarial Networks~\cite{gan2014goodfellow}
and Variational Auto-Encoders (VAE)~\cite{kingma2013auto} are widely use to synthesize realistic images from a latent code. 
In contrast, our goal is to perform conditional image synthesis which allows more fine-grained control over the image generation process.
Some popular conditional image generation approaches are label-to-image translation~\cite{odena2017conditional,miyato2018cgans}, image-to-image translation~\cite{pix2pix2017isola,dosovitskiy2017learning,crn2017chen,cyclegan2017zhu,unit2017liu,pix2pixhd2018wang} and text-to-image generation~\cite{reed2016generative,zhang2017stackgan,hong2018inferring,xu2018attngan}. Earlier works~\cite{pix2pix2017isola,dosovitskiy2017learning,crn2017chen} on conditional image-to-image generation are mostly supervised with paired data from both domains. Several works~\cite{cyclegan2017zhu,unit2017liu} propose a way to use unpaired data from both domains for conditional image generation. 
Other advances in conditional image generation include innovations in network architectures and loss functions for generating high resolution images~\cite{pix2pixhd2018wang} and generating multiple diverse images~\cite{choi2018stargan,munit2018huang,yang2017diversity,hui2018unsupervised}. In this work, we develop a model for photo-realistic geometry-to-image translation using only unpaired training data as supervision.
Our work is closely related to~\cite{alhaija2018geometric} which also considers geometry-to-image translation, but requires paired training data.
Our work belongs to the family of unpaired conditional image generation models with architecture and losses (e.g., shared discriminator) specialized for the geometry-to-image translation.
Our model outperforms state-of-the-art unpaired image-to-image translation models \cite{cyclegan2017zhu,munit2018huang} by a large margin.

\noindent \textbf{Intrinsic Image Decomposition} is a long standing problem in computer vision. \cite{barron_sfs}  poses the task as an optimization problem with a set of hand-crafted priors for shape, shading albedo etc. On the other hand supervised methods like \cite{sfsnetSengupta18, janner2017intrinsic, neuralSengupta19} use synthetic data to train the model followed by refinement on real images. However, synthetic data might not capture all the real world statistics and models trained with synthetic data might now generalize well to real images. Recently, several self-supervised intrinsic decomposition methods have been proposed \cite{BigTimeLi18, wei-chu-iid, liu2020usi3d}. \cite{liu2020usi3d} uses single images during both training and inference stages. \cite{BigTimeLi18, wei-chu-iid} make use of multiview images or video sequences of the scene during training and infer on a single image . Our work falls in the realm of self-supervised intrinsic image decomposition. We do not use any paired synthetic data or multiview sequences to train out model. Instead, we rely on jointly training models for neural rendering and image decomposition.

\vspace{-0.2cm}
\section{Method} \label{sec:method}
\begin{figure*}[t!]
\centering
    \includegraphics[width=0.95\textwidth]{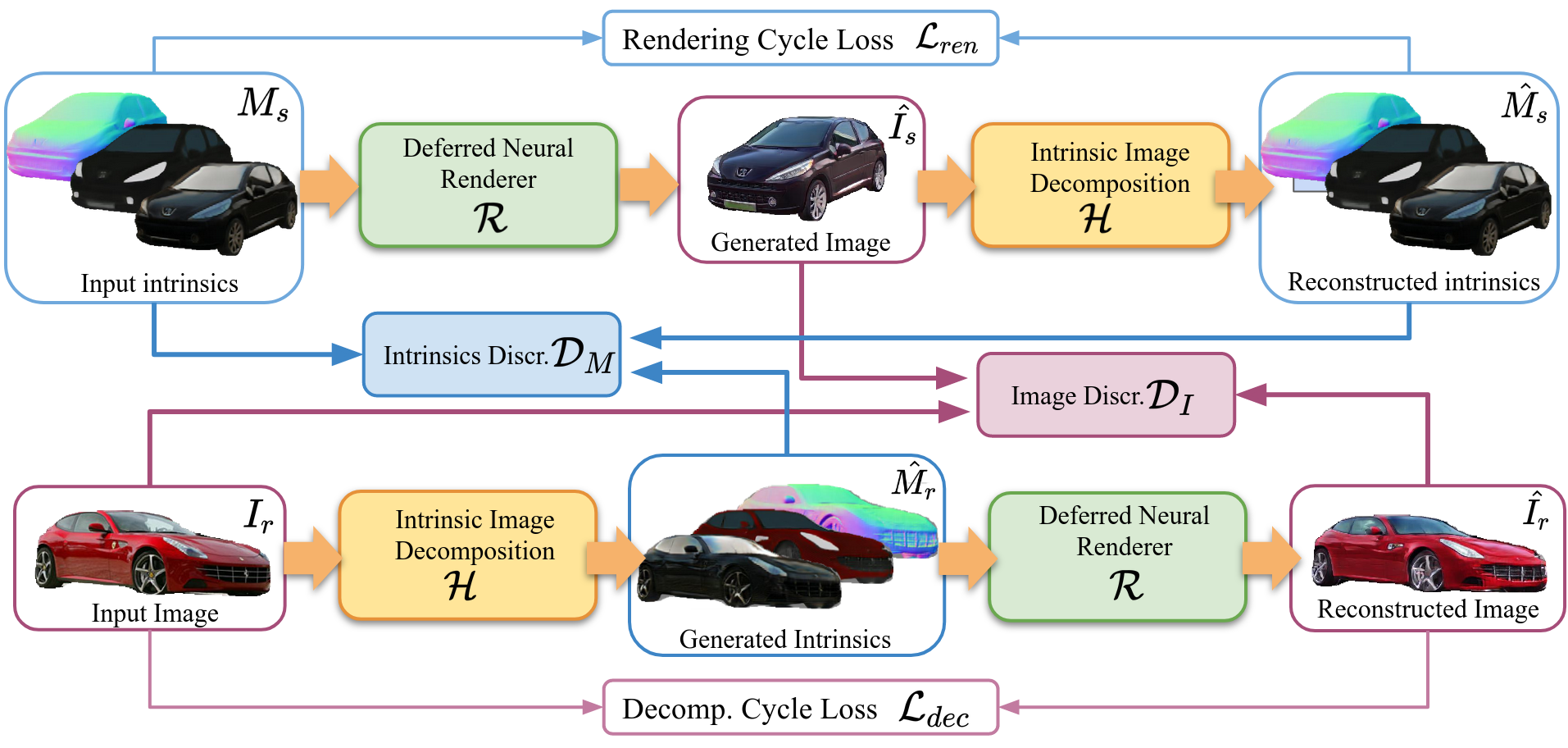}
    \caption{\textbf{Intrinsic Autoencoder.} Our model comprises two cycles: The first cycle (blue) auto-encodes a set of intrinsics rendered from 3D CAD models using appearance as latent representation. The second cycle (red) auto-encodes real images using image intrinsics as representation. Consistency is achieved through a combination of cycle losses and shared adversarial losses. Networks sharing the same weights are illustrated with the same color (green/yellow).}
\label{fig:method}
\end{figure*}

Our Intrinsic Autoencoder model (Fig.~\ref{fig:method}) consists of two generator networks $\mathcal{R}$ and $\mathcal{H}$ for Deferred Neural Rendering and Intrinsic Image Decomposition, respectively. 
The Deferred Neural Rendering Network $\mathcal{R}: M \rightarrow \hat{I}$ takes as input a set of intrinsic maps $M=\{A,N,F\}$. %
The object's surface normal vectors in the view coordinate system $N\in \mathbb{R}^{ H \times W \times 3}$ provides the Deferred Neural Rendering Network important information about the local shape of the object which is necessary for creating shading and reflection in the output image. 
The albedo $A\in \mathbb{R}^{ H \times W \times 3}$ is a pixel-wise RGB value that describes the material or texture color at every pixel, ignoring any lighting effects.
Finally, the environment reflections $F\in \mathbb{R}^{ H \times W \times 3}$ are computed by projecting a high dynamic range environment map onto the 3D model.
Note that this simple projection operation does not involve any complicated sampling or ray-tracing operations. 
As shown in our experiments, the Deferred Neural Renderer can also be trained with a subset of those inputs since it is able to compensate for the missing information. 
The DNR network $\mathcal{R}: M \rightarrow \hat{I}$ transforms all the input intrinsics $M$ into a realistic image $\hat{I} \in \mathbb{R}^{ H \times W \times 3} $ that corresponds to the input intrinsics.
Similarly, the Intrinsic Image Decomposition (IID) network $\mathcal{H}: I \rightarrow \hat{M}$ performs the opposite task by taking an input image $I$ and predicting its intrinsics $\hat{M} \in \mathbb{R}^{ H \times W \times 9}$.\\ 
Supervised training of $\mathcal{R}$ and $\mathcal{H}$ on real data is typically difficult due to the lack of real training image and intrinsics pairs $(I_r,M_r)$. 
Instead, we use a combination of cycle-consistency losses and adversarial losses that require no paired training examples. 
This allows us to leverage a large dataset of real images $\{I_r^i\}_{i=0}^{n}$ and an unpaired set of synthetically generated intrinsic maps $\{M_s^i\}_{i=0}^{m}$.  
In the following, we detail our cycle consistency losses and the novel shared adversarial losses.

\vspace{-0.2cm}
\subsection{Cycle Consistency}\label{sec:cyc_cons}
\vspace{-0.1cm}
\noindent \textbf{Rendering Cycle.}
The goal of the rendering cycle is to train $\mathcal{R}$ in order to produce realistic images $\hat{I}_s=\mathcal{R}(M_s)$ from synthetic intrinsic maps $M_s$. 
To train $\mathcal{R}$ without paired data, we use the inverse transformation $\mathcal{H}$ which decomposes the predicted image $\hat{I}_s$ back into its intrinsic maps $\hat{M}_s = \mathcal{H}(\mathcal{R}(M_s))$ as illustrated in Fig.~\ref{fig:method}. We encourage consistency of the intrinsics using the rendering cycle consistency loss which is defined as the \textit{Smooth-$L_1$} distance between the input and reconstructed intrinsics 
\begin{equation} 
  \mathcal{L}_{ren}(\mathcal{R},\mathcal{H},M_s) = \|\mathcal{H}(\mathcal{R}(M_s)) - M_s\|_1.
\end{equation}
\noindent \textbf{Decomposition Cycle.}
Similarly, we train $\mathcal{H}$ to generate intrinsic maps $\hat{M}_r = \mathcal{H}(I_r)$ from real images $I_r$. 
To ensure consistency with the input $I_r$, the output intrinsics $\hat{M}_r$ are passed to the deferred neural renderer $\mathcal{R}$ to reconstruct the image $\hat{I}_r = \mathcal{R}(\mathcal{H}(I_r))$. The decomposition cycle consistency loss is defined by:
\begin{equation}
  \mathcal{L}_{dec}(\mathcal{R},\mathcal{H},I_r) = \|\mathcal{R}(\mathcal{H}(I_r)) - I_r\|_1.
\end{equation} 
The combined cycle consistency loss is then defined as: 
\begin{equation}
  \mathcal{L}_{cyc}(\mathcal{R},\mathcal{H},I_r,M_s) =
  \mathcal{L}_{ren}(\mathcal{R},\mathcal{H},M_s) +
  \mathcal{L}_{dec}(\mathcal{R},\mathcal{H},I_r) 
\end{equation}
To ensure that the predicted normals $\hat{N}_s = \mathcal{H}_N(I_r) $ and the reconstructed real normals $\hat{N}_r = \mathcal{H}_N(\mathcal{R}(M_s)) $ are properly normalized, we exploit an additional normalization loss $\mathcal{L}_{\text{norm}}$:
$$\mathcal{L}_{\text{norm}}(\mathcal{R},\mathcal{H}, I) =  \mid 1-{\|\mathcal{H}_N(I_r)\|}_2\mid + \mid 1-{\|\mathcal{H}_N(\mathcal{R}(M_s))\|}_2\mid$$ 

\subsection{Shared Adversarial Loss}\label{sec:shared_discr}
\vspace{-0.2cm}
While the cycle consistency loss ensures that the network input can be reconstructed from its output, it doesn't place any importance on the realism of that output. 
Additionally, the cycle consistency loss assumes a one-to-one deterministic mapping between the input and output.
While this is a reasonable condition for some image-to-image translation tasks \cite{cyclegan2017zhu}, it is violated when translating between images and their intrinsic properties.
Decomposing an RGB image into its high-dimensional intrinsic properties is a one-to-many transformation since multiple decompositions can be consistent at the same time with the same image, e.g., a gray patch may correspond to a gray diffuse surface or a black glossy surface with specular highlight. 
Likewise, the process of creating an image from an incomplete set of intrinsic properties involves making additional predictions about missing attributes like lighting conditions, optical aberrations, noise or higher-order light interactions.
To better capture this multi-modal relationship, we use an adversarial loss between the input and its reconstruction.  

An adversarial discriminator $\mathcal{D}$ is a classification model trained to predict if a data sample is produced by a generative model or if it stems from the true data  distribution.  
To train our Intrinsic Autoencoder, we use two adversarial discriminators, $\mathcal{D}_I$ for discriminating generated images $\hat{I}_{\{s,r\}}$ from real images $I_{r}$, and $\mathcal{D}_M$ for discriminating generated intrinsic maps $\hat{M}_{\{r,s\}}$ from synthetic intrinsic maps $M_{s}$.
The discriminators help our model to learn the distribution of real images and synthetic intrinsics by optimizing the following adversarial \cite{gan2014goodfellow} loss function
\small
\begin{equation}
  \mathcal{L}_{\text{adv}}(\mathcal{R},\mathcal{H},\mathcal{D}_I,\mathcal{D}_M) = \mathcal{L}^I_{\text{adv}}(\mathcal{R},\mathcal{H},\mathcal{D}_I) + \mathcal{L}^M_{\text{adv}}(\mathcal{R},\mathcal{H},\mathcal{D}_M)
\end{equation}
\normalsize
where
\vspace{-0.2cm}
\begin{equation}
\small
\begin{aligned}
   \label{eq:loss_adv_I}
     \mathcal{L}^I_{\text{adv}}(\mathcal{R},\mathcal{H},\mathcal{D}_I) &=\log(\mathcal{D}_I(I_{r})) + \log (1-\mathcal{D}_I(\mathcal{R}(M_s)) \\
      &+ \log (1-\mathcal{D}_I(\mathcal{R}(\mathcal{H}(I_r)))
\end{aligned}
\end{equation}

\normalsize
is our novel \textit{shared adversarial image loss} which discriminates both between the real image $I_r$ and the generated synthetic image $\hat{I}_s=\mathcal{R}(M_s)$, as well as between the real image $I_r$ and the reconstructed real image $\hat{I}_r = \mathcal{R}(\mathcal{H}(I_r))$.
Similarly, we define the \textit{shared adversarial intrinsic loss} as
\small
\begin{equation}
\begin{aligned}
   \label{eq:loss_adv_M}
  \mathcal{L}^M_{\text{adv}}(\mathcal{R},\mathcal{H},\mathcal{D}_M) &= \log(\mathcal{D_M}(M_s)) + \mathbb\log (1-\mathcal{D}_M(\mathcal{H}(I_r)) \\
  &+ \log(1-\mathcal{D}_M(\mathcal{H}(\mathcal{R}(M_s))))
\end{aligned}
\end{equation}
\normalsize
Using the reconstructed inputs $\hat{I}_r$ and $\hat{M}_s$ in addition to the generated samples $\hat{I}_s$ and $\hat{M}_r$ for training $\mathcal{D}_I$ and $\mathcal{D}_M$ makes the discriminators more robust
and prevents overfitting.
This is especially important when a relatively small number of 3D objects are used to create the synthetic intrinsic maps which can lead to a discriminator that recognizes the model features rather that the image realism. 

\vspace{-0.1cm}
\subsection{Implementation and Training}
\vspace{-0.1cm}
We train our Intrinsic Autoencoder networks $\mathcal{R}, \mathcal{H}$ in addition to the adversarial discriminators $\mathcal{D}_M, \mathcal{D}_I$ from scratch by optimizing the joint objective 
\vspace{-0.1cm}
\begin{equation}
  \label{eq:objective_function}
  \min_{\mathcal{R,H}}\max_{\mathcal{D}_I,\mathcal{D}_M} \mathcal{L}_{\text{cyc}} + \mathcal{L}_{\text{norm}} + \mathcal{L}_{\text{adv}}
\end{equation}

Our framework is implemented in PyTorch \cite{pytorch} and trained using Adam \cite{KingmaB14} with a learning rate of 0.0002.
The Deferred Neural Rendering Network is a coarse-to-fine generator introduced in \cite{pix2pixhd2018wang} for the deferred neural rendering network. The input to the network is of size $256\times512\time9$ constructed by concatenating normals, albedo and reflections. The output of the network is an RGB image of size $256\times512\times3$.

We use three networks $\mathcal{H}$ = \{$\mathcal{H}_N$, $\mathcal{H}_A$, $\mathcal{H}_F$\} for estimating the surface normals $N$, Albedo $A$ and environment reflections $F$, respectively, from an image $I$.
Each network has a ResNet architecture with 5 ResNet blocks.\\
\noindent \textbf{Adversarial Discriminator Networks.}
Since the local structure of the generated images is mostly controlled by the input intrinsics, we want the image discriminator $\mathcal{D}_I$ to mainly focus on the global realism of the output.  
To address this, we use a multi-scale PatchGAN \cite{pix2pixhd2018wang} discriminator which comprises two fully-convolutional networks that classify the local image patches at two scales, full and half resolution. The discriminator outputs a realism score for each patch instead of a single prediction per image. This has been shown to produce more detailed images for similar conditional image generation tasks \cite{pix2pix2017isola,pix2pixhd2018wang,cyclegan2017zhu}. The intrinsics discriminator $\mathcal{D}_M$ has the same architecture except that the input is a 9-channel stack combining all three intrinsic maps. We found that using a single discriminator for the combination of the intrinsic maps performs better than separate networks for each. This is likely due to the inter-dependence between the different intrinsic properties that allows the discriminator to detect inconsistencies between the generated intrinsic maps. We provide more architecture and training details in the supplementary material.

\vspace{-0.2cm}
\section{Experiments}
\label{sec:exp}
\noindent \textbf{Synthetic Data Generation.} \label{sec:synth_training_data}
To generate the synthetic training data, we use dataset from \cite{Alhaija2018IJCV} containing 28 3D car models covering 6 car categories (SUV, sedan, hatchback, station wagon, mini-van and van). Apart from the geometry, we do not need any physically-based materials or textures for the models. Instead, we assign to each car part a simple material with only two properties, the color and a scalar glossiness factor for computing reflection maps. 
We assign each 3D car part a fixed material from a set of 18 fixed materials. Additionally, we randomly pick one of 15 materials with different colors for the car body during the rendering process.
Next, a camera position is randomly chosen within a radius of 8 meters and a maximum height of 3 meters. We use a fast OpenGL based rendering engine which operates at around 3 frames per second including the model loading time. It outputs the surface normals of the car model in the camera coordinate space and the albedo channels indicating the material color at each pixel without any lighting or shading. Finally, we produce the environmental reflections by using a 360 degree environment map from \cite{Alhaija2018IJCV}. These kind of reflections are very efficient to compute since they only require the view vector and the surface normal and do not rely on expensive path-tracing. We render 20,000 synthetic samples of normals, albedo and reflections.%

\noindent \textbf{Real training data.} \label{sec:real_training_data}
We obtain the real images from a fine grained car classification dataset presented in \cite{KrauseStarkDengFei-Fei_3DRR2013}. For convenience, we refer to this as the real car dataset. It contains 16,000 images of cars captured in various lighting conditions, resolutions and poses and with different camera sensors and lenses. 

\subsection{Baselines} \label{sec:baselines}
Since our goal is to train with only unpaired data, we choose to benchmark our method against two state-of-the-art unpaired image generation approaches,  CycleGAN\cite{cyclegan2017zhu} and MUNIT\cite{munit2018huang}. However, since both methods were originally designed for image-to-image translation rather than deferred rendering, we setup two additional strong baselines that highlight the importance of our contributions in improving the quality of our results. \\  
\noindent \textbf{CycleGAN and MUNIT.} CycleGAN\cite{cyclegan2017zhu} is a generic method for translating between two domains without available paired data. MUNIT\cite{munit2018huang} aims at producing a diverse set of translations between different domains. We modify the two methods slightly to use our stacked 9 channel synthetic intrinsic maps as inputs.
\\
\noindent \textbf{Without shared discriminator.}
In this setup, we do not use the shared adversarial discriminator discussed in \ref{sec:shared_discr}. Instead, we only use the discriminator $\mathcal{D}_I$ between generated image $\hat{I}_s$, real image ${I}_r$. Similarly, the discriminator $\mathcal{D}_M$ is used only between synthetic intrinsics ${M}_s$ and generated intrinsics $\hat{M}_r$. \\
\noindent \textbf{Only rendering cycle.}
Here, we train the model using only the deferred rendering cycle discussed in (Sec. \ref{sec:cyc_cons}) and do not use the decomposition cycle.\\

\begin{figure*}[th!]
	\centering
	\includegraphics[width=\textwidth]{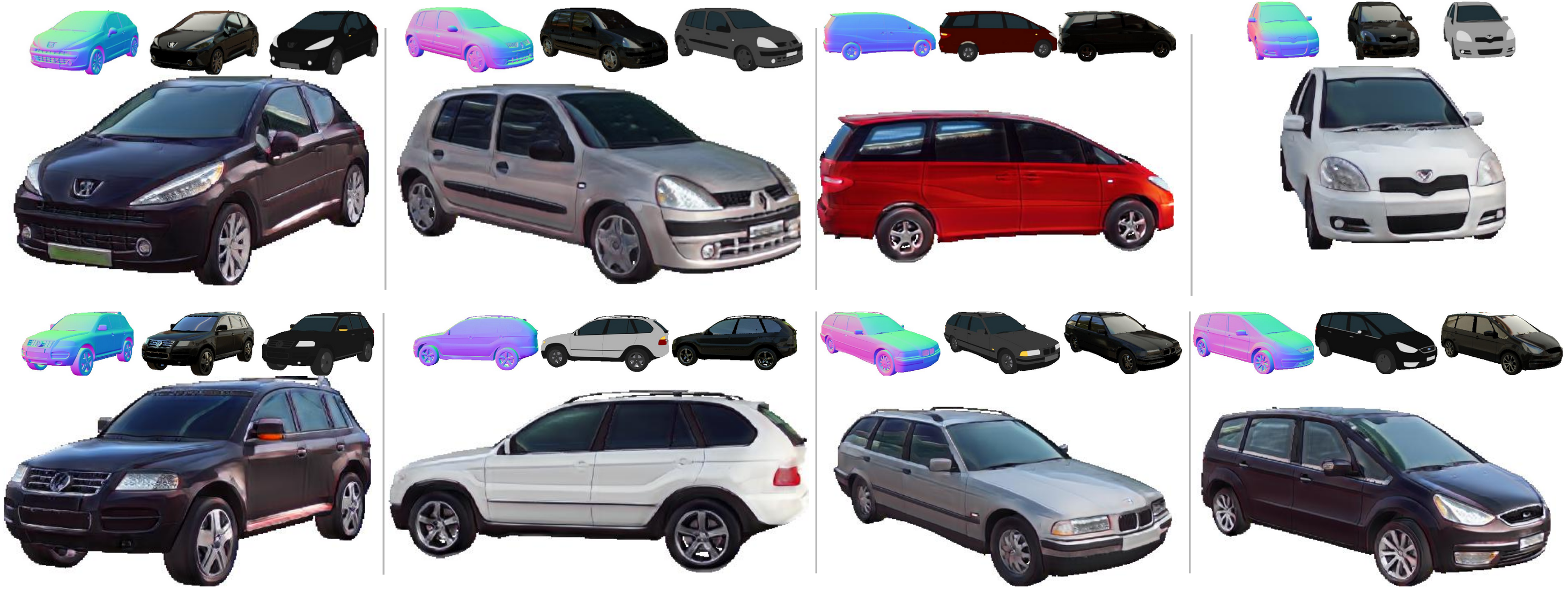}
	\caption{\textbf{Images generated using our Deferred Neural Renderer.} Inputs to the network are intrinsic maps consisting of albedo, normals and reflections, shown above the generated images.}
	\label{fig:results_render}
\end{figure*}

\begin{figure*}[th!]
	\centering
	\includegraphics[width=\textwidth]{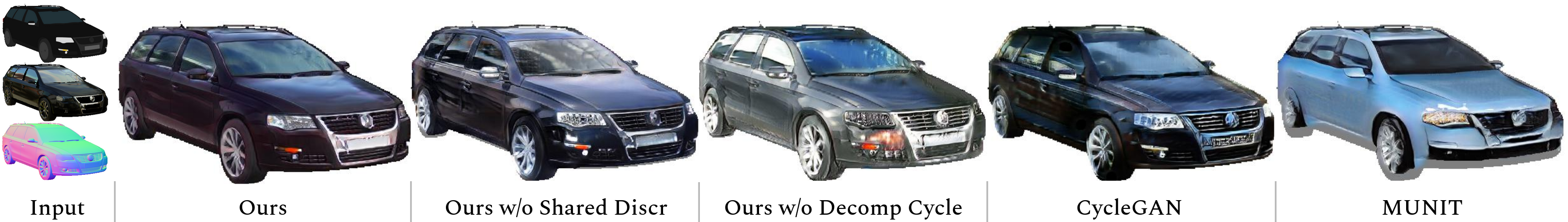}
	\caption{\textbf{Qualitative Comparison with baselines on Neural Rendering.} Inputs to the network are intrinsic maps consisting of albedo, normals and reflections, shown above the generated images. Additional higher resolution results are provided in the supplementary materials.}
	\label{fig:qual_comparison}
\end{figure*}

\vspace{-0.3cm}
\subsection{Deferred Neural Rendering} \label{sec:def_ren}
To evaluate our approach for deferred neural rendering, we use the network $\mathcal{R}$ to produce images given synthetic intrinsic maps (albedo, normals, reflections) and  compare it to other baselines, both qualitatively and quantitatively.
\vspace{-0.3cm}
\subsubsection{Qualitative results}
Fig.~\ref{fig:results_render} shows car images generated using our deferred neural renderer from the input synthetic intrinsic maps shown above them. The car models in the evaluation set have been previously seen by the generator, but the unique combination of pose and paint color has not been seen during training. Our approach is able to generate detailed photo-realistic images of cars with consistent geometry and distinct parts. We emphasize that the deferred neural rendering network is trained without any rendered or real geometry-image pairs. Instead, it is able to learn the appearance of different car parts from a large set of real car images. \\ 
In Fig.~\ref{fig:qual_comparison} we compare the results of our full model to various baselines. The results clearly show the improvements in visual quality achieved when using our full model. Specifically, MUNIT appears to be unable to preserve the geometry and albedo of the input in the generated image,  CycleGAN images has significant artefacts on the windows, body, etc. When training our model without the shared discriminator, the resulting images suffer from irregular reflection patterns and a noisy image. This is likely due to the strong overfitting required by the network to reproduce the input image exactly when using only an $L_1$ loss. The model trained without the decomposition cycle is not able to preserve the input intrinsics in terms of albedo and reflection.

\begin{figure}[ht!]
	\includegraphics[width=\columnwidth]{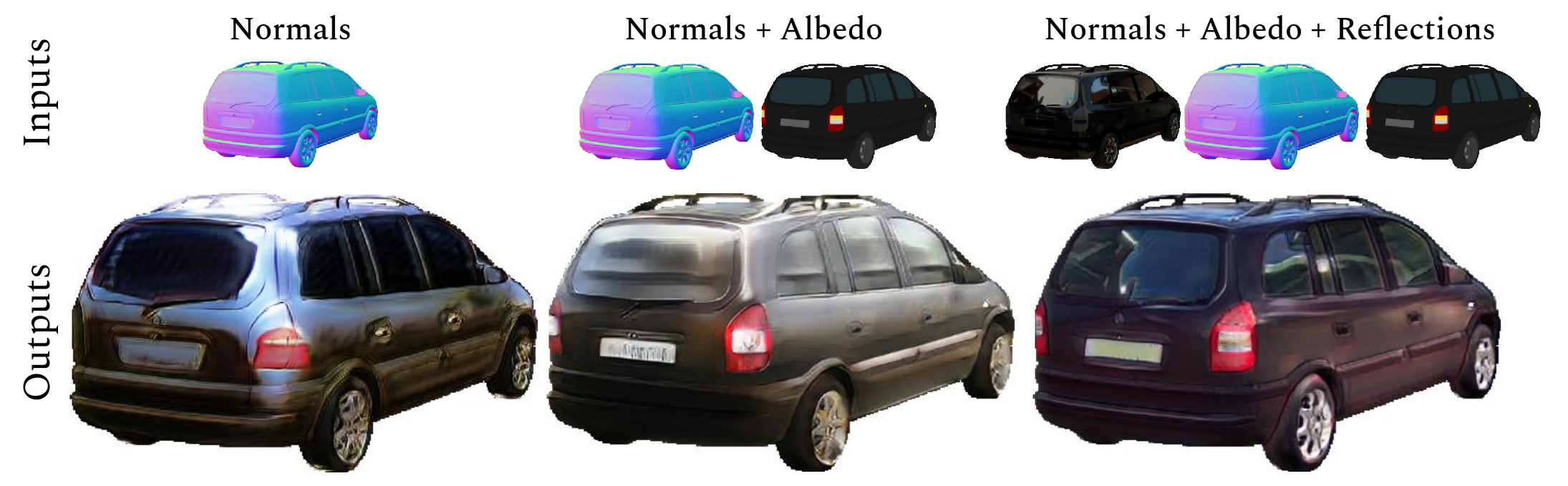}
	\caption{ \textbf{Images generated using models trained with ablated inputs.}	
	}
	\vspace{-0.3cm}
	\label{fig:input_albation}
\end{figure}

In figure \ref{fig:input_albation}, we show the effect of input intrinsic maps on the quality of rendered images. When the model is trained only with normals as intrinsic input, the geometry of the result is well rendered but the color of different parts poorly defined. The model trained on both normals and albedo demonstrates sharper image quality  but the hallucinated reflections by the network lacks lack realistic details. Finally, using the environmental reflections helps the network produce consistent and realistic images with sharp details. 

\vspace{-0.2cm}
\subsubsection{Quantitative results}
\vspace{-0.2cm}
We evaluate the quality of generated images using Fr\'echet Inception Distance(FID) \cite{fid2017heusel} and Kernel Inception Distance(KID) \cite{binkowski2018demystifying}. Both metrics compute the distance between the features of two sets of images, obtained from a pre-trained CNN. 
Table \ref{tab:fid_scores} presents both the FID and KID between the images generated using various methods and the real images. Our full model achieves the lowest FID and KID values (47.6, 4.2) indicating that the rendered images from our model are closest to the distribution of real images compared to MUNIT \cite{munit2018huang} and CycleGAN \cite{cyclegan2017zhu}. 
Further, when we ablate each of the intrinsic map inputs, both FID and KID increase substantially. Notably, in the case of ablating albedo input, the highest increase in distances can be observed (88.7, 5.4), implying its importance for photo-realistic image generation. 
We conclude that albedo is the most important for our task followed by normals and reflections maps. In both cases where we ablate the decomposition cycle or rendering cycle, we observe a huge increase in the distances signifying the importance of using both cycle consistency losses during training. Finally, training with the setup of separate discriminators as mentioned in \ref{sec:baselines} leads to an increase in the distances.

\setlength{\tabcolsep}{2pt} 
\begin{table}\centering
	\begin{tabular}{l|cc|c|cc|ccc}
		\toprule
		 &       	  &        			&    		   & w/     & w/o    &     &     & \\
		 & Cycle      &   	   			&  			   & Shared & Decom. & w/o & w/o & w/o\\
		 & GAN        & \small{MUNIT}  & \textbf{Ours} & Discr  & Cyc. 	 & $A$ & $N$ & $F$\\
		      
		\midrule
		FID &  103.3  	      &  99.0   &  \textbf{47.6}  & 59.2  & 99.6   & 88.7  & 60.2  & 56.7 \\[2pt]
		KID &   10.2  	 	  &  13.5   &  \textbf{4.2}   &  4.8  &  11.8  &  5.4  &  4.9  &  5.9 \\
		\midrule
	\end{tabular}
\caption{ \textbf{FID and KID between real images and generated samples.} All inputs are provided to the generator (Albedo, Normals and Reflections).}
\label{tab:fid_scores}
\end{table}

\vspace{-0.3cm}
\subsubsection{Human Experiments}
\vspace{-0.2cm}
We design two experiments to measure the visual realism of generated car images using the Amazon Mechanical Turk platform to crowd source human evaluations. 
For each comparison, we presented 40 human subjects each with 50 image pairs to choose the more realistic looking image. The results are presented in Table \ref{tab:human_exps}. The first row presents experiments where one image is picked from the real images and the other is from one of the synthesis methods and presented in a random order. Images from our full model seem to be most confused with real car image since only 67.5\% of choices were correct while in 32.5\% of the trials the subjects choose our images to be the real one.\\
In the second experiment subjects are presented with an image generated by our full model and a matching image generated by one other synthesis methods. The results in the second row of Table \ref{tab:human_exps} show that subjects choose our results to be more realistic over $80\%$ of times when compare to CycleGAN and MUNIT. This clearly indicates a high level of visual quality of our generated images compared to those generated from existing methods. On the other hand, images from our ablated models appear to be much closer to our full model visual quality. 

\setlength{\tabcolsep}{2pt} 
\begin{table}\centering
	\begin{tabular}{l|cc|c|cc}
		\toprule
		                & Cycle      &        &  			  & w/o Shared & w/o Decom \\
		                & GAN        & MUNIT  & \textbf{Ours} & Discr. & Cyc. \\
		\midrule
		Real Im &    77.7\%	      &  75.6\%   &       \textbf{67.5\%}   &     68.9\%     &   71.0\%  \\[2pt]
		Ours &    80.0\% 	  &     85.8\% &  --   & 57.6\%   &    63.8\% \\
		\midrule
	\end{tabular}
\vspace{-0.2cm}
\caption{\textbf{Human Subject Study.} Comparisons to identify realistic images in an A/B test using Amazon Mechanical Turk. The numbers indicate the ratio of trials where the image from real or our model was chosen as more realistic compared to the image from the method on the header.}
\label{tab:human_exps}
\vspace{-0.3cm}
\end{table}

\subsection{Intrinsic Image Decomposition} \label{sec:img_dec}

\vspace{-0.1cm}
\subsubsection{Qualitative results}
\begin{figure*}[th!]
	\centering
	\includegraphics[width=\textwidth]{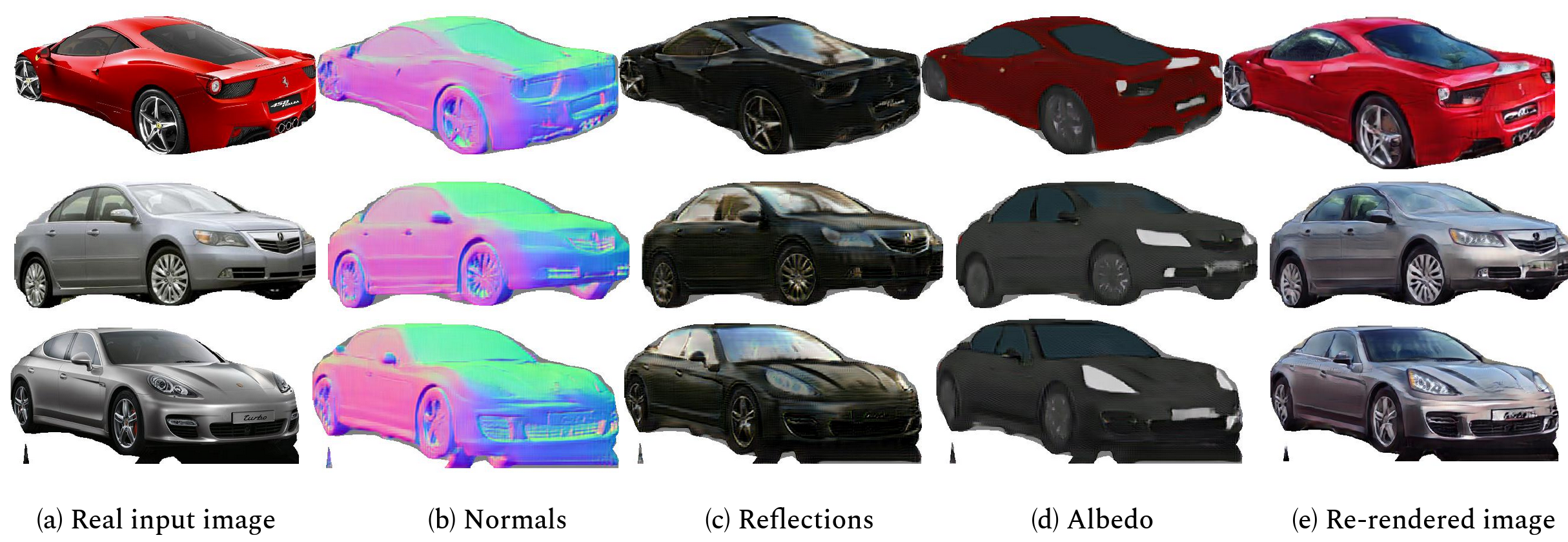}
	\caption{\textbf{Results of our intrinsic decomposition network on real images.} The first column shows the inputs to the network. Our model is able to decompose the sport car in first row accurately even though our synthetic training dataset doesn't include any sport cars at all. The car models of other inputs images are also not present in our synthetic dataset. } 
	\label{fig:results_decomp}
\end{figure*}

\vspace{-0.2cm}
In fig. \ref{fig:results_decomp}, we show that the intrinsic decomposition network is able to decompose real car images into their intrinsic maps. We would like to emphasize that the model does not have access to ground-truth intrinsic maps for real images during the training phase. Also, these car models are not present in the synthetic training data.

\begin{figure*}[th!]
	\centering
	\includegraphics[width=\textwidth]{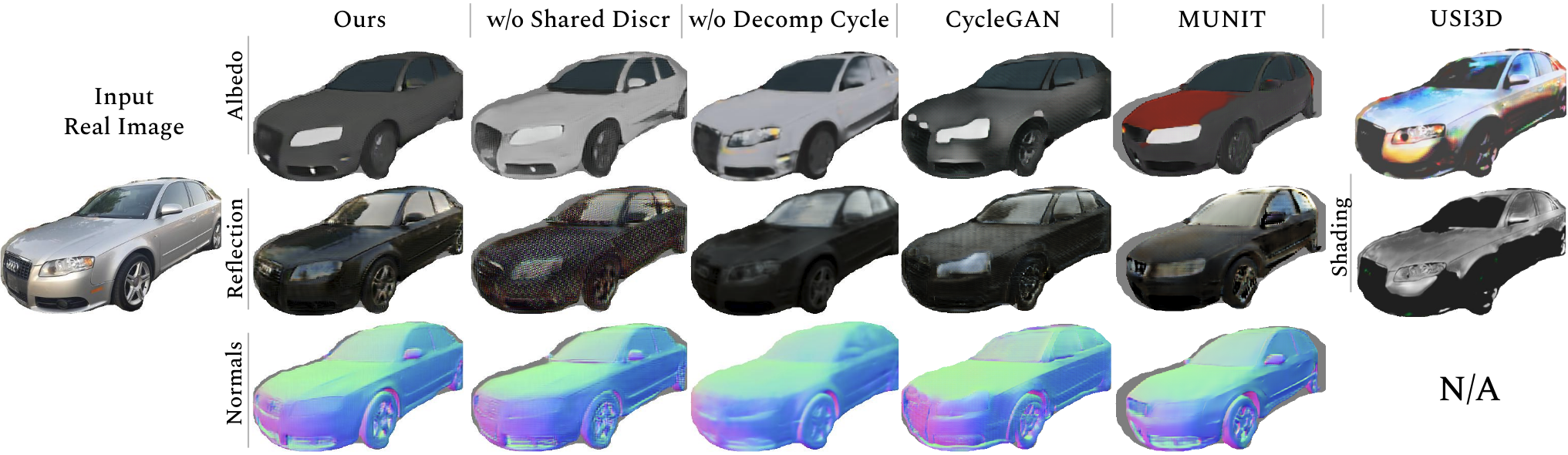}
	\caption{\textbf{Comparison with Baselines for intrinsic decomposition.}
	Note that USI3D \cite{liu2020usi3d} only produces albedo and shading and not reflections or normals.}
	\label{fig:results_compare_decomp}
	\vspace{-0.2cm}
\end{figure*}

Figure \ref{fig:results_compare_decomp} compares the decompositions produced by our model to those from other baselines. 
Both CycleGAN \cite{cyclegan2017zhu} and MUNIT \cite{munit2018huang} show significant artificats and inconsistencies when trained to decompose real images.
The USI3D \cite{liu2020usi3d} fails to generalize to real models since it was trained using synthetic data from ShapeNet \cite{shapenet2015}. 
Our model without decomposition cycle also recovers noisy albedo and normals due to  overfit only to synthetic data. On the other hand, training without the shared discriminator leads to severe artefacts. This is because the rendering network tries to encode intrinsics information in the generated images in the form of high frequency artefacts such that the decomposition network can easily recover them. 

\vspace{-0.2cm}
\subsubsection{Quantitative results}
\vspace{-0.2cm}
To evaluate the intrinsic maps predicted by the intrinsic image decomposition network ($\mathcal{H}$) we construct a synthetic dataset containing rendered RGB images and their corresponding intrinsic maps rendered using a standard Physically Based Renderer (Blender \cite{Blender}). To obtain the error between predicted and ground truth normals, we compute the average cosine distance between them. The errors for albedo and reflection are the average $\ell_1$ distances between the predicted and ground truth maps. Table \ref{tab:decomp_errors} presents the errors of various methods for predicting intrinsic maps. 
Our full model has the least error for all the modalities followed by our model without the shared discriminator, without decomposition cycle and finally MUNIT and CycleGAN. This indicates that our model is able to learn accurate image decomposition while keeping generalization. Note that these PBR-rendered images have not been presented to our network during training.

\setlength{\tabcolsep}{1pt} 
\begin{table}\centering
	\begin{tabular}{llcccccc}
		\toprule
		
		&						& \multicolumn{1}{c}{w/o Shared}    & \multicolumn{1}{c}{w/o Decom.} & \multicolumn{1}{c}{Cycle}	& 	& \\
		&						& \multicolumn{1}{c}{Discr.}    & \multicolumn{1}{c}{cycle} & \multicolumn{1}{c}{GAN}	& \multicolumn{1}{c}{MUNIT}	& Ours\\

		\midrule
		\rule{0pt}{1ex}
		&Normal Err.              &   17.75$^\circ$    &  18.80$^\circ$   &  27.82$^\circ$  & 29.15$^\circ$ & \textbf{14.73$^\circ$} \\[2pt]
		&Albedo Err.               &   54.00     &  67.21  &  68.18  & 81.44 & \textbf{52.74}  \\[2pt]
		&Reflection Err.               &   55.60  &  71.00  &  73.18  & 74.75 & \textbf{51.74}  \\
		\midrule
	\end{tabular}
	\caption{ \textbf{Errors for the Intrinsic Decomposition Task.} Our method achieves the lowest error on all tasks.}
	\label{tab:decomp_errors}
\end{table}

\vspace{-0.1cm}
\subsection{Results on ShapeNet Aeroplanes}
\begin{figure*}[th!]
	\centering
	\includegraphics[width=\textwidth]{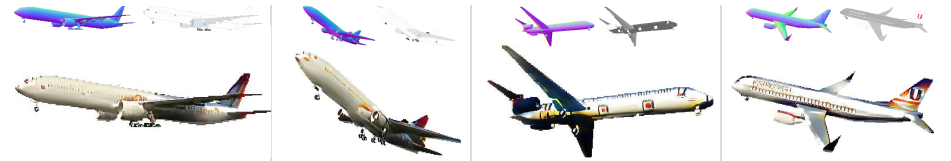}
	\caption{ \textbf{Images generated by our network trained on aeroplanes from ShapeNet \cite{shapenet2015}.}	
}
	\label{fig:aeroplanes}
	\vspace{-0.3cm}
\end{figure*}

\vspace{-0.1cm}
We train our model for the object class "Aeroplanes". We obtain the real images from FGVC-Aircraft dataset \cite{maji13fine-grained} which contains 10,000 images of aeroplanes. We use the 3D models of aeroplanes from the Shapenet dataset \cite{shapenet2015} to obtain our intrinsic maps. We follow the process mentioned in sec.\ref{sec:synth_training_data} to generate input training data. We use the normals and albedo as inputs to the network. Figure \ref{fig:aeroplanes} illustrates realistic images generated using our deferred rendering network, demonstrating the ability of our method to handle low-quality mesh and texture models.

\section{Conclusion}
In this paper, we presented a joint approach for training a deferred rendering network for generating realistic images from synthetic image intrinsics and an intrinsic image decomposition network for decomposing real images of an object into its intrinsic properties. We trained the model using unpaired 3D models and real images. Our qualitative and quantitative experiments revealed that using a combination of shared adversarial losses and cycle consistency losses is able to produce images that are both realistic and consistent with the control input.

\section{Acknowledgement}
\noindent
This project has received funding from European Research Council (ERC) under the European Unions Horizon 2020 programme (grant No. 647769) and Heidelberg Collaboratory for Image Processing (HCI).
{\small
\bibliographystyle{ieee}
\bibliography{bibliography_custom}
}

\end{document}